\documentclass[letterpaper, 10 pt, conference]{ieeeconf}

\IEEEoverridecommandlockouts
\usepackage{cite}
\usepackage{amsmath,amssymb,amsfonts}
\usepackage{algorithm, algorithmic}
\usepackage{graphicx}
\usepackage{textcomp}
\usepackage{xcolor}
\usepackage{svg}
\usepackage{array}
\usepackage{booktabs}
\usepackage{optidef}

\usepackage{pgfplots}
\pgfplotsset{compat=newest}
\usepackage{tikzscale}

\usepackage{hyperref}

\title{\LARGE \bf
A Dynamics Simulator for Soft Growing Robots*
}

\author{
Rianna Jitosho$^{1}$, Nathaniel Agharese$^{1}$, Allison Okamura$^{1}$, and Zac Manchester$^{2}$
\thanks{*This material is based upon work supported in part by the National Science Foundation (NSF), including the Graduate Research Fellowship Program and Awards 1637446 and 2024247, and the Air Force Office of Scientific Research under Award FA2386-17-1-4658.}
\thanks{$^{1}$R. Jitosho, N. Agharese, and A. M. Okamura are with the Department of Mechanical Engineering, Stanford University, Stanford, CA 94305 USA.}%
\thanks{$^{2}$Z. Manchester is with the Robotics Institute, Carnegie Mellon University, Pittsburgh, PA 15213 USA.}%
\thanks{Corresponding author e-mail: rjitosho@stanford.edu}%
}

\begin{document}

\maketitle
\thispagestyle{empty}
\pagestyle{empty}

\begin{abstract}
Simulating soft robots in cluttered environments remains an open problem due to the challenge of capturing complex dynamics and interactions with the environment. Furthermore, fast simulation is desired for quickly exploring robot behaviors in the context of motion planning. In this paper, we examine a particular class of inflated-beam soft growing robots called ``vine robots", and present a dynamics simulator that captures general behaviors, handles robot-object interactions, and runs faster than real time. The simulator framework uses a simplified multi-link, rigid-body model with contact constraints. To narrow the sim-to-real gap, we develop methods for fitting model parameters based on video data of a robot in motion and in contact with an environment. We provide examples of simulations, including several with fit parameters, to show the qualitative and quantitative agreement between simulated and real behaviors. Our work demonstrates the capabilities of this high-speed dynamics simulator and its potential for use in the control of soft robots.
\end{abstract}

\newcommand{\T}{^{\top}}
\newcommand{\qkp}{q_{k+1}}
\newcommand{\vkp}{v_{k+1}}

\section{Introduction}
Dynamics simulators are widely used in robotics for applications such as design exploration, motion planning, controller tuning, and human operator training. However, simulating soft robots in cluttered environments remains an open problem due to the challenge of modeling complex dynamics and handling environment interactions. There are simulators that use Finite Element Analysis to achieve high accuracy \cite{Faure2012}, but they are often slow and difficult to use. In this work, we are interested in a simulator that is suitable for motion planning, in which it is common to opt for a low-fidelity model in order to compute dynamics quickly. Thus, we can define application-specific goals. In particular, we are interested in a dynamics simulator that: (1) captures general behaviors, (2) handles robot-object interactions, (3) runs faster than real time, and (4) is amenable to classical motion planning strategies.

\begin{figure}[t]
    \centering
    \includegraphics[width = .48\textwidth]{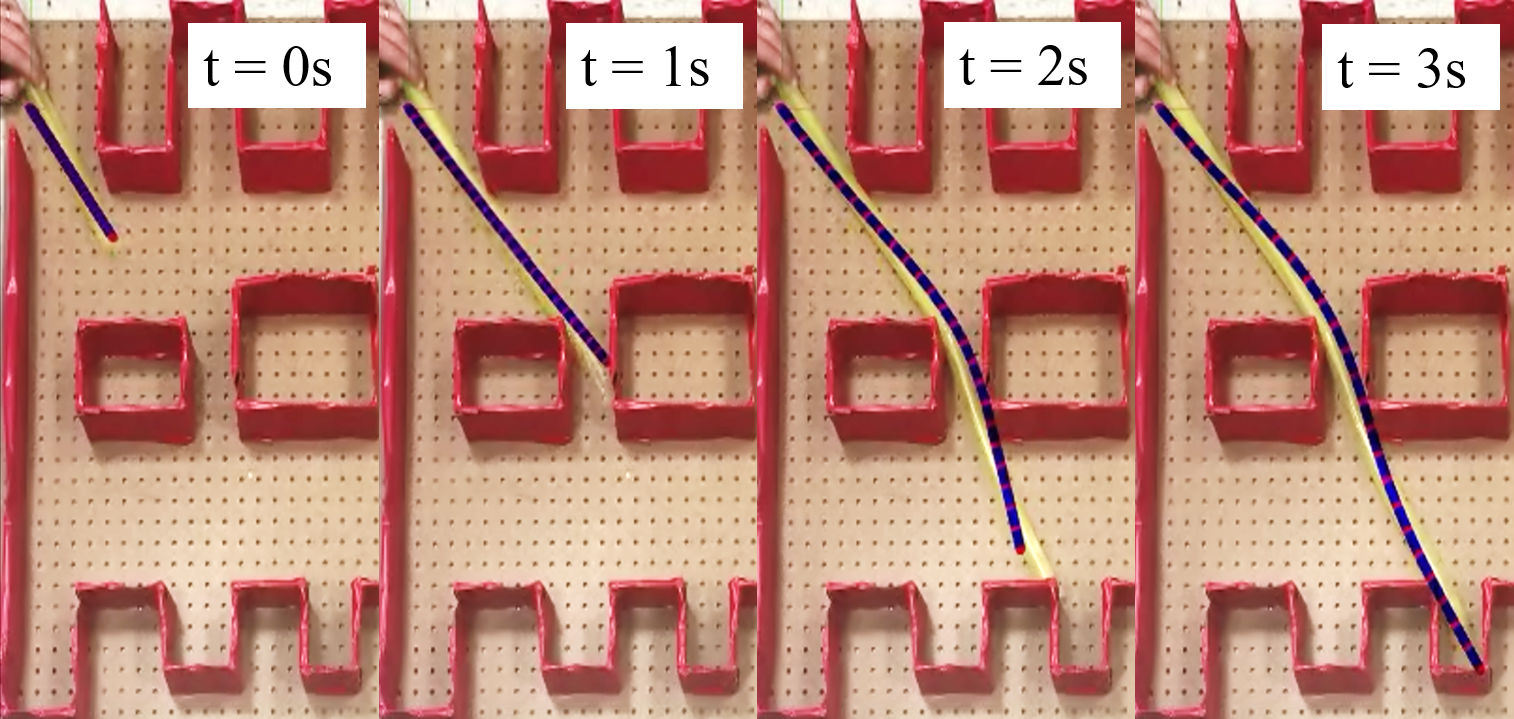}
    \caption{Trajectory of a simulated robot (blue) overlaid on video frames of a soft growing robot (yellow) in a cluttered environment. The simulated trajectory was produced with the framework presented in this paper. This simulator captures general robot behaviors, handles environment interaction, and runs in real time.}
    \label{fig:Overview}
\end{figure}

Real-time simulators currently exist for elastomer-based soft robots \cite{Hu2019,Huang2020}. In contrast, our work examines a particular class of soft growing robots called ``vine robots,'' which has shown potential for use in applications such as medical devices as well as navigation and exploration\cite{Slade2017,Coad2020}. Unlike elastomer-based robots, these inflated-beam robots are built from materials that feature flexibility but little to no extensibility. These robots can grow over 100 times their original length via pneumatic eversion. In Fig.\ \ref{fig:Overview}, we provide an example of a vine robot in motion, and overlay a simulated trajectory that was generated using the framework presented in this paper. 

Prior work has contributed models characterizing vine robot behavior \cite{Hawkes2017,Blumenschein2018,Blumenschein2017,Greer2018}, and recent papers have explored models of vine robot interactions with the environment. Haggerty, et al.\ modeled transverse and axial buckling modes that result from environment interaction to predict the pressure required for eversion \cite{Haggerty2019}. Greer, et al.\ developed a kinematic model for a vine robot growing in a cluttered environment \cite{Greer2020}. Selvaggio, et al.\ examined robot-obstacle interactions in the context of workspace analysis and presented an object-interaction model that accounts for point contacts \cite{Selvaggio2020}. These models are for planar motion, as is the case in our work.

The differences between our work and existing literature are as follows: First, we present a simplified rigid-body dynamic model that can handle interaction with objects of various geometries. This is more general than the existing kinematic and object-interaction models. Second, we incorporate this dynamic model into a simulator framework. Vine robot growth has previously been simulated in Unity \cite{El-Hussieny2018}, but the growth was simulated by incrementally adding rigid bodies. This requires a state vector that increases in length over time, which is not amenable to common motion planning strategies. In contrast, the dynamic model presented in this paper has a fixed-length state vector. 

The main contributions of our work are: 

\begin{enumerate}
    \item A vine robot simulator framework that captures general behaviors, handles robot-object interactions, runs faster than real time, and is amenable to motion planning.
    \item A method for fitting model parameters based on video data of a vine robot in motion.
    \item Validation of the framework by comparing simulations to real vine robot behaviors.
\end{enumerate} 

The remainder of the paper is organized as follows: Section II describes the details of the simulator framework. Section III presents a method for fitting model parameters to reduce the sim-to-real gap. Section IV characterizes the performance of the simulator framework in order to validate the methods proposed in Sections II and III. Finally, Section~V summarizes our conclusions.

\section{Simulator Framework}\label{sec:framework}

We draw from previous work on dynamic simulation of rigid bodies with contact to design a simulator that is computationally efficient. We make three key design choices: First, we use a dynamic model with a constant-size state vector so that computational complexity remains fixed regardless of robot growth over time. Moreover, constant-size state vectors are amenable to classical motion planning strategies. Second, we use an impulse-velocity dynamics formulation. Because collisions between rigid bodies result in velocity discontinuities, instantaneous accelerations and forces at the moment of impact become infinite. To address this, dynamics are written in terms of velocities and impulses. Third, we use the Lagrange multiplier formulation originally presented by Baraff \cite{Baraff1996} to define the model and equations of motion. The benefits of the Lagrange multiplier formulation are in its simplicity: (1) all coordinates are in the world frame, and (2) no recursive computations are required (as is the case with the recursive Newton-Euler and Featherstone algorithms \cite{Luh1980, Featherstone2008}). Furthermore, Baraff \cite{Baraff1996} demonstrates that these benefits can be achieved while matching the computational efficiency of traditional minimal-coordinate techniques. 

In the remainder of this section, we present the details of the simulator framework. In \ref{subsec:lagrange}, we provide additional background information on the Lagrange multiplier method. In \ref{subsec:model}, we explain the simplified rigid-body model used to abstract the vine robot. In \ref{subsec:joint} to \ref{subsec:growth}, we describe the constraints that must be satisfied at any given time. In \ref{subsec:stiff_damp} and \ref{subsec:dynamics}, we write out the stiffness and damping models as well as the full robot dynamic model. Finally, in \ref{subsec:sim_algo}, we combine all previous subsections into a single algorithm for simulating vine robot dynamics.

\subsection{Lagrange Multiplier Formulation}\label{subsec:lagrange}

In the Lagrange multiplier formulation, rigid bodies are described with maximal coordinates, which equates to 3 degrees of freedom for each rigid body (assuming planar motion). Joints between bodies are modeled with explicit constraints, and Lagrange multipliers associated with constraint forces (or in this work, constraint impulses) are calculated at each time step along with the next state. 

Constraints are written in the form $c(q) = 0$, where the function $c(q)$ defines a vector of constraint errors given a robot configuration $q$. The constraint impulses act in opposition to the constraint errors. Specifically, the constraint impulse vector $\lambda$ is defined in a particular basis that depends on $c(q)$, and the mapping of $\lambda$ from the constraint basis to maximal coordinates is the Jacobian of the constraint equation, i.e. $J = \frac{\partial c}{\partial q}(q)$. This yields the following dynamics:
\begin{align} \label{eq:lagrange_dynamics}
    M\Delta v = J \T \lambda + F \Delta t, 
\end{align}
where $M$ is the (constant) mass matrix, $\Delta v = \vkp - v_k$ is the change in velocity from timestep $k$ to $k+1$, $J\T$ maps constraint impulses $\lambda$ into maximal coordinates, $F$ contains other forces such as gravity, and $\Delta t$ is the timestep length. 

We note two important characteristics of the dynamics formulation. First, dynamics are at the level of velocities and impulses rather than accelerations and forces due to the discontinuities that arise in contact dynamics. Second, this is an implicit formulation wherein simulating dynamics requires finding $\lambda$ and $\vkp$ such that the resulting motion satisfies the system constraints at timestep $k+1$. Thus, this formulation does not suffer from the constraint drift problem common in naive maximal-coordinate implementations.

\subsection{Rigid-Body Model}\label{subsec:model}

\begin{figure}[bp]
    \centering
    \includegraphics[width = .41\textwidth]{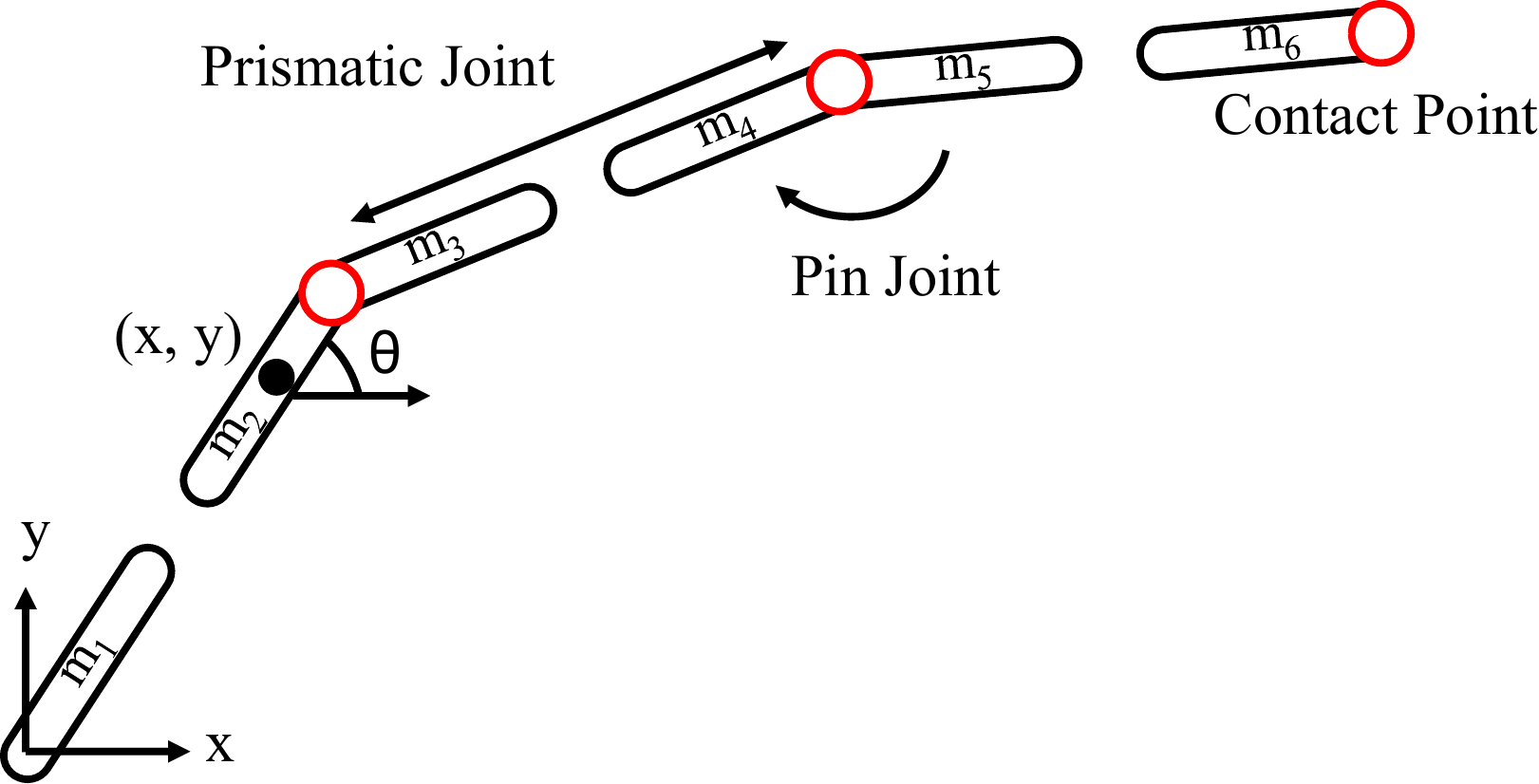}
    \caption{Illustration of the rigid-body model used in the simulator. The vine robot is abstracted as a series of rigid bodies ($m_1, \dots, m_n$) with alternating pin and prismatic joints. For each rigid body, the configuration contains the position of the center of mass and the orientation in the world frame. Illustrated in red are the contact points, which are used for handling environment interactions.}
    \label{fig:state_vector}
\end{figure}

We define a dynamic model that abstracts the vine robot as a series of rigid bodies. Since we use maximal coordinates, the configuration for each body is $(x, y, \theta)$, where $(x, y)$ is the position of the center of mass and $\theta$ is the orientation, all expressed in the world frame (Fig.\ \ref{fig:state_vector}). The configuration $q$ for the entire vine robot is the concatenation of the configurations for each body, ordered from the base to the tip. Similarly, the velocity vector $v$ for the entire vine robot is the concatenation of the velocity vectors for each body.

The series of rigid bodies are connected via alternating pin and prismatic joints. The first pin joint is between the first body (at the base of the robot) and the world frame origin. An illustration of these alternating joints is shown in Fig.\ \ref{fig:state_vector}. The use of prismatic joints is an abstraction of the actual mechanism for growth. In reality, vine robots grow via eversion, i.e. lengthening is achieved by adding new material to the tip. However, our abstraction enables a fixed-length state vector, which is important for fast computation and compatibility with motion planning.

\subsection{Joint Constraints}\label{subsec:joint}

In this section, we define the joint constraints of the rigid-body model.

\subsubsection{Pin Joint}

\begin{figure}[bp]
    \centering
    \includegraphics[height=2.6cm]{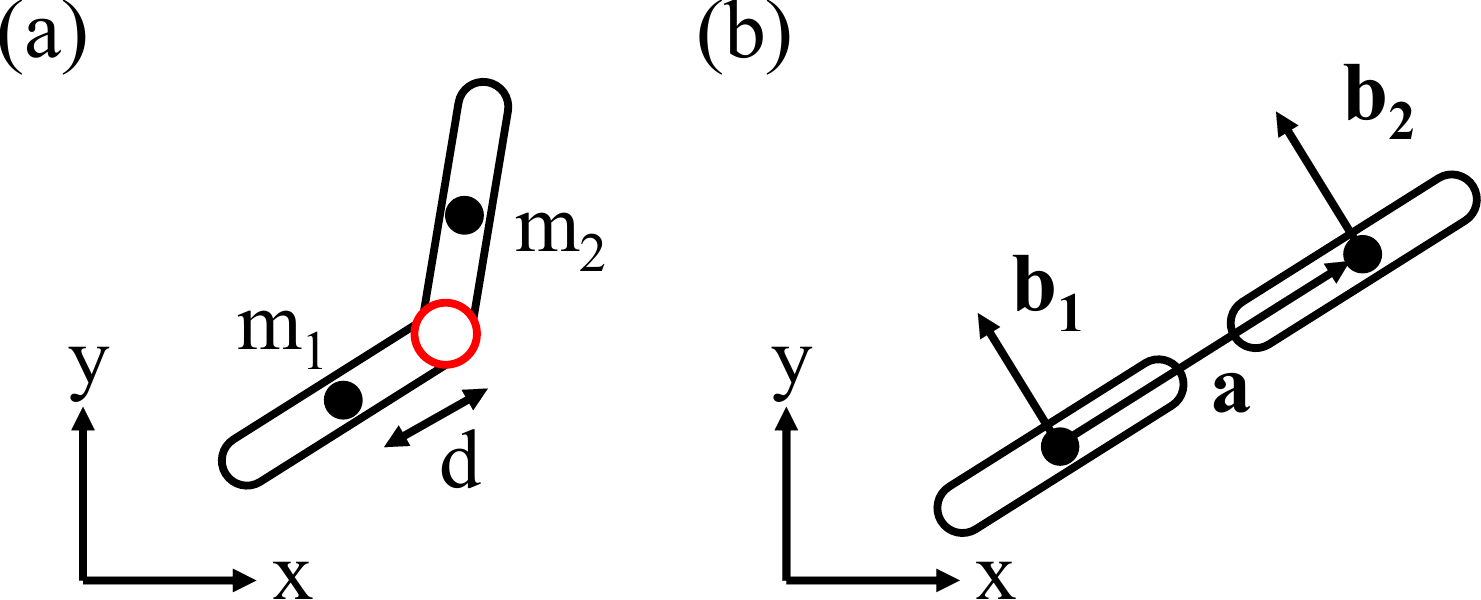}
    \caption{Illustrations of joint constraints. (a) For a pin joint, endpoints from body 1 ($m_1$) and body 2 ($m_2$) must be coincident (red circle). Endpoint coordinates are computed with the configuration of each body and the distance $d$ from the center of mass to the endpoint. (b) For a prismatic joint, we define normal vectors $b_1$ and $b_2$ for body 1 and 2 respectively, as well as vector $a$, which goes from body 1's center of mass to that of body 2. The prismatic joint constraint is that the two bodies be collinear, i.e. $\mathbf{b_1} \cdot \mathbf{a} = \mathbf{b_2} \cdot \mathbf{a} = 0$.}
    \label{fig:joint_constraint}
\end{figure}

Let bodies~1 and 2 be connected via a pin joint, where body~2 is more distal. The corresponding pin joint constraint is that the distal endpoint of body~1 must have the same coordinates as the proximal endpoint of body~2 as illustrated in Fig.\ \ref{fig:joint_constraint}. In algebraic form, the pin constraint is the vector equation: 
\begin{equation} \label{eq:pin-constraint}
    \begin{bmatrix} 
    (x_1 + d \cos(\theta_1)) - (x_2 - d \cos(\theta_2)) \\ 
    (y_1 + d \sin(\theta_1)) - (y_2 - d \cos(\theta_2)) 
    \end{bmatrix}
    = \mathbf{0},
\end{equation}
where $(x_i, y_i, \theta_i)$ is the configuration of body $i$ and $d$ is the distance from the center of mass to the body's endpoint.

\subsubsection{Prismatic Joint}

Let bodies 1 and 2 be connected via a prismatic joint, where body 2 is more distal. The corresponding prismatic joint constraint is that the two bodies must be collinear. For this to be true, the following vector equation must be satisfied:
\begin{equation} \label{eq:prismatic-constraint}
    \begin{bmatrix} 
    \mathbf{b_1} \cdot \mathbf{a}\\ 
    \mathbf{b_2} \cdot \mathbf{a} 
    \end{bmatrix}
    = \mathbf{0},
\end{equation}
where $\mathbf{b_i}$ is the vector normal to body $i$ and $\mathbf{a}$ is the vector from body 1's center of mass to that of body 2 (Fig.\ \ref{fig:joint_constraint}). 

\subsubsection{Total Joint Constraint Equation}

The joint constraint for the entire robot is the concatenation of Equations \eqref{eq:pin-constraint} and \eqref{eq:prismatic-constraint} for all bodies:
\begin{equation}\label{eq:joint_con}
    c(q) = 0,
\end{equation}
where $c(q)$ computes the left-hand sides of the joint constraint equations for each joint given robot configuration $q$.

\subsection{Contact Constraints}

We enforce contact constraints at specific ``contact points" along the robot. The contact points in Figs.\ \ref{fig:state_vector} and \ref{fig:joint_constraint} are shown in red. In our implementation, contact points correspond to pin joints, but this choice is arbitrary, and any general design choice for number and location of contact points is allowed.

Each contact point must satisfy the contact complementary constraints, which can be summarized as: (1) no interpenetration (a contact point cannot be inside an object); (2) normal forces exerted by objects must be non-negative (an object cannot pull the robot towards it); and (3) normal forces can only be exerted when the distance between the contact point and the object is zero, 
\begin{equation}\label{contact_con}
\begin{aligned}
    \Phi(q) &\geq 0\\
    n &\geq 0\\
    \Phi(q) \T n &= 0,
\end{aligned}
\end{equation}
where $\Phi$ computes the vector of signed distances between each contact point and each object in the environment, and $n$ is a vector of the normal forces exerted by the environment at the contact points. The direction of each normal force is the outward-pointing surface normal vector of each object, so $n$ need only store a scalar value for each normal force.

This method for handling contact interactions is an approximation of actual contact interactions whereby contact can occur at any point along the robot. Potential problems arise when the spacing between contact points is large enough that simulations become unrealistic (e.g.~an object can pass through the robot). Thus, it is important to place contact points such that the distance between neighboring points is small compared to the size of objects in the environment. This results in a trade-off where increasing the number of contact points improves the accuracy of contact behavior but increases computational complexity. As a rule of thumb, the maximum distance between contact points should be smaller than the minimum corner radius of all objects.

\subsection{Growth Constraints}\label{subsec:growth}

In our framework, we treat the growth rate of the robot as a control input $u$. In doing so, we introduce a growth rate constraint and a corresponding constraint impulse $w$. While we could have defined growth forces and included this in the external force vector $F$ in Equation (\ref{eq:lagrange_dynamics}), we chose to add a growth rate constraint because prior work with hardware implementations often used a feedback controller for the growth rate \cite{Coad2020,Luong2019}. With a feedback controller, actuation is adjusted to reach a target growth rate. Similarly, with the growth rate constraint in our simulator framework, constraint impulses are computed to achieve a target growth rate. The growth constraint is represented as:

\begin{equation}
    g(q, v) - u = 0 \label{eq:growth_con},
\end{equation}
where the function $g(q, v)$ computes the growth rate of each prismatic joint via kinematics given the configuration $q$ and velocity $v$ of the robot.

\subsection{Bending Stiffness and Damping} \label{subsec:stiff_damp}

We model the system response to bending with torsion springs and dampers at the pin joints. We express this as:

\begin{equation}\label{eq:spring_damper}
    \tau (q,v) = -K \theta(q) - C \Dot{\theta} (v),
\end{equation}
where $K$ is the stiffness matrix, $\theta(q)$ computes the pin joint angles given a robot configuration, $C$ is the damping matrix, and $\Dot{\theta}(v)$ computes the pin joint velocities given a robot velocity. Incorporating this into the robot dynamics given by Equation (\ref{eq:lagrange_dynamics}) requires a matrix $R$ that maps these torques into maximal coordinates:
\begin{equation}\label{eq:R_tau}
    F = R\tau(q,v).
\end{equation}

\subsection{Dynamics}\label{subsec:dynamics}

Having defined the system constraints and external forces, we can write out the robot dynamics. The system constraints include joint constraints, contact constraints, and growth constraints. Per the Lagrange multiplier formulation, we introduce three corresponding constraint impulses that we refer to as $\lambda$, $n$, and $w$. Combining these with the spring and damping forces yields:
\begin{equation}\label{eq:dyn_con}
M(\vkp - v_k) = J\T \lambda + L\T n + G\T w + F\Delta t, \end{equation}
where $M$ is the mass matrix, $v_i$ is the velocity at timestep $i$, $\lambda$, $n$, and $w$ are constraint impulses, $J\T$, $L\T$, and $G\T$ are Jacobians that map the constraint impulses into maximal coordinates, $F$ captures spring and damping torques, and $\Delta t$ is the length of the timestep. 

For the configuration dynamics, we use the semi-implicit Euler method commonly used for simulating rigid-body dynamics with contact \cite{Stewart1996}. Specifically,
\begin{equation} \label{eq:semi_implicit_euler}
\qkp = q_k + \vkp \Delta t,
\end{equation}
where $q_k$ and $\qkp$ are the configurations at timestep $k$ and $k+1$ respectively.

\subsection{Simulator Algorithm}\label{subsec:sim_algo}

We now synthesize the information from the previous sections and present the simulator algorithm. The high-level steps are listed in Algorithm \ref{alg:simulate}.

\begin{algorithm}[bp]
\caption{Simulate Dynamics}
\begin{algorithmic}
\label{alg:simulate}
\REQUIRE $q_k$, $v_k$, $u_k$, $\Delta t$
\STATE compute $J(q_k)$ and $c(q_k)$
\STATE compute $L(q_k)$ and $\Phi(q_k)$
\STATE compute $G(q_k, v_k)$ and $g(q_k, v_k)$
\STATE solve Quadratic Program (QP)
\STATE $\vkp \leftarrow$ QP optimal solution
\STATE $\qkp \leftarrow q_k + \vkp \Delta t$
\RETURN $\qkp$, $\vkp$
\end{algorithmic}
\end{algorithm}

Simulating the dynamics requires finding values for the constraint impulses such that the resulting motion satisfies the system constraints at the end of the time step. This is equivalent to solving a feasibility problem satisfying: 
\begin{equation}
\begin{aligned}
M(\vkp - v_k) &= J\T \lambda + L\T n + G\T w + F\Delta t \\
\qkp &= q_k + \vkp \Delta t\\
c(\qkp) &= 0\\
\Phi(\qkp) &\geq 0\\
n &\geq 0\\
\Phi(\qkp) \T n &= 0\\
g(\qkp, \vkp) - u &= 0, 
\end{aligned}
\end{equation}
where the decision variables are $\qkp$, $\vkp$, $\lambda$, $n$, and $w$.

We apply two simplifications that reduce the computation cost. First, we linearize the constraints and use Equation (\ref{eq:semi_implicit_euler}) to approximate $\delta{q}$ as $\vkp \Delta t$. For the joint constraint,
\begin{equation}
    c(\qkp) \approx c(q_k) + J\vkp \Delta t,
\end{equation}
where $J = \frac{\partial c}{\partial q}(q_k)$.
For the contact constraint,
\begin{equation}
    \Phi(\qkp) \approx \Phi(q_k) + L \vkp \Delta t,
\end{equation}
where $L = \frac{\partial \Phi}{\partial q}(q_k)$. 
For the growth constraint,
\begin{align}
\begin{split}
    g(\qkp, \vkp) &\approx g(q_k, v_k) + g_q \vkp \Delta t + g_v\Delta v\\
    &\approx \Bar{g}(q_k, v_k) + \Bar{G}\vkp,
\end{split}
\end{align}
where 
\begin{align*}
g_q &= \frac{\partial g}{\partial q}(q_k),\\
g_v &= \frac{\partial g}{\partial v}(v_k),\\
\Delta v &= \vkp - v_k,\\
\Bar{g}(q_k, v_k) &= g(q_k, v_k) - v_k, \text{and}\\
\Bar{G} &= g_q \vkp \Delta t + g_v.
\end{align*}


These linearizations eliminate the dependence of the constraints on $\qkp$. Thus, we remove $\qkp$ from the feasibility problem and compute it with Equation (\ref{eq:semi_implicit_euler}) once $\vkp$ is known. 
Second, we reformulate the linearized feasibility problem as the following quadratic program (QP): 
\begin{mini}
  {\vkp}{\frac{1}{2}v_{k+1} \T M v_{k+1} - v_{k+1} \T (Mv_k + F\Delta t)}{}{}
  \addConstraint{c(q_k) + Jv_{k+1}\Delta t}{=0}{}
  \addConstraint{\Phi (q_k) + Lv_{k+1}\Delta t}{\geq 0}{}
  \addConstraint{\Bar{g}(q_k, v_k) + \Bar{G}v_{k+1} - u_k}{=0.}{}
\end{mini}

The feasibility problem corresponds to the Karush-Kuhn-Tucker (KKT) conditions of this QP. This allows us to leverage fast, efficient QP solvers, and in this work we use the Operator Splitting Quadratic Program solver\cite{Stellato2020}.

\section{Fitting Model Parameters}\label{sec:fitting}

The framework presented in Section \ref{sec:framework} is sufficient for generating qualitatively realistic vine robot behaviors. However, it is necessary to bridge the sim-to-real gap by fitting model parameters to data. In our work, the rigid-body model is an abstraction of the vine robot, so we are unable to collect explicit measurements of the state vector defined in Section \ref{subsec:model}. Thus, we are faced with a joint state \emph{and} parameter estimation problem. A common technique for estimating parameters given missing or hidden variables is the Expectation-Maximization algorithm (EM) \cite{Kokkala2014}. However, we can collect video data of a vine robot in motion, which provides an observation of the hidden state variables. By leveraging this additional information, we can use a batch nonlinear least-squares method and avoid complexities from the recursion in EM. 

We pose the joint state and parameter estimation problem as a single optimization problem and simultaneously fit model parameters as well as a trajectory. This method assumes the presence of errors in both the dynamic model as well as the measurements (i.e. video data). Thus, rather than treating the measurements as ground truth and placing hard constraints on satisfying dynamics, we form an objective function with two goals: (1) minimize deviations from the reference trajectory, and (2) minimize deviations from the dynamic model. Specifically, we solve:

\begin{mini}
  {Z, K, C, r}{\sum_{i=1}^{N}{||\hat{p}_i - p(q_i)||^2} + \sum_{i=1}^{N-1}{r_i\T Rr_i}}{}{}
  \addConstraint{Z_{i+1}-f(Z_i, K, C)} {=r_i,\quad}{i=0,\ldots,N-1}
  \addConstraint{c(q_i)}{=0,\quad}{i=0,\ldots,N}
  \addConstraint{g(q_i, v_i)}{=u_i,\quad}{i=0,\ldots,N}
  \addConstraint{\phi(q_i)}{=0,\quad}{i=0,\ldots,N}.
  \label{eq:fit}
\end{mini}
where $Z = \{Z_1, \dots, Z_N \}$ is a robot trajectory over $N$ timesteps, $Z_i$ is the concatenation of configuration $q_i$ and velocity $v_i$, $K$ and $C$ are stiffness and damping matrices as defined in Equation (\ref{eq:spring_damper}), and $r_i$ is the deviation from the dynamic model from timestep $i$ to $i+1$. Measurements $\hat{p}_i = \{p_i^1, \dots, p_i^n\}$ are points that are evenly spaced along the length of the real robot in the $i^{th}$ video frame. We use computer vision techniques to find these coordinates for each frame in the video. In order to compare this set of points with a simulated vine configuration $q_i$, we define a function $p(q)$ that interpolates $n$ points along the length of the simulated robot in order to correspond with the extracted measurements $\hat{p}_i$. $R$ is a diagonal matrix of weights that roughly translates to the level of trust in the dynamic model. This is because increasing the values in $R$ will encourage lower values for $r$ and thus smaller deviations from the model. $f(Z_i, K, C)$ computes the state at the next time step. The last three constraints ensure that the joint, contact, and growth constraints are satisfied at all timesteps. Because we have a video reference, we know \textit{a priori} which rigid bodies are in contact with an object at each time step. Thus, we use a modified contact constraint $\phi(q_i)$ to compute the signed distances for cases where the distance ought to be 0.

\section{Performance}

In this section, we characterize the performance of our simulator framework. First, we validate runtime performance and measure how this scales with model size. Second, we discuss simulator accuracy by comparing real vine robot behaviors with simulated behaviors produced with fit models. Finally, we demonstrate general environment interactions by providing an example simulation in a cluttered environment. 

\subsection{Runtime and Scaling} \label{subsec:runtime}

\begin{figure}[tp]
    \centering
    \includegraphics[width=.35\textwidth]{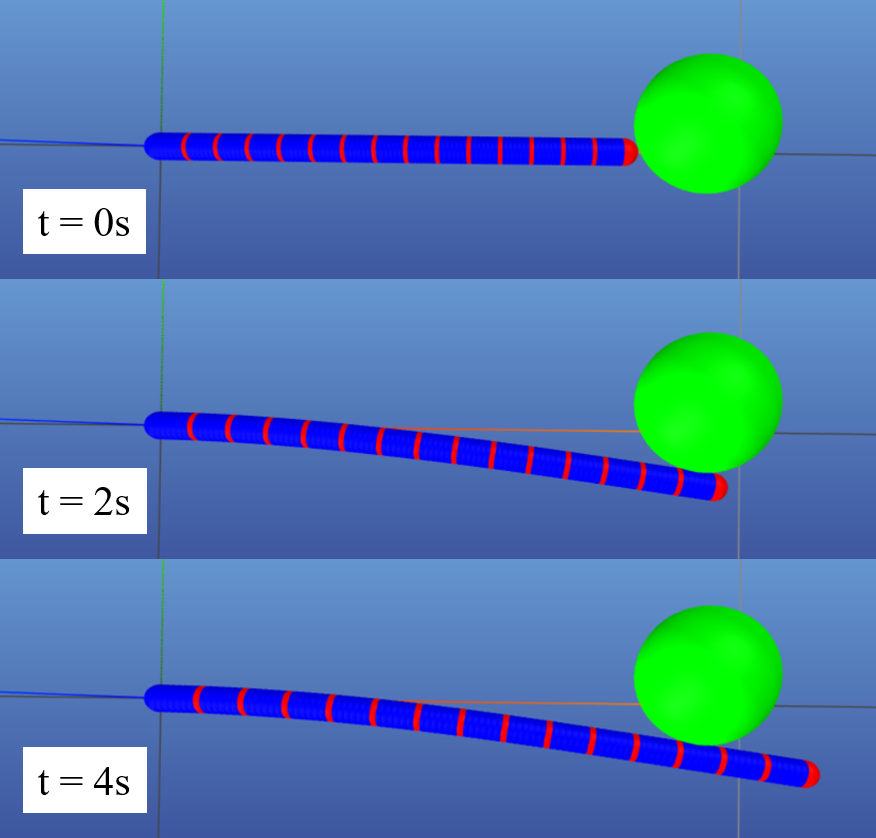}
    \caption{Frames from a visualization of a simulated trajectory. A circular object in the environment is shown in green. The left end of the robot was fixed to the origin, and a constant growth rate control input was applied. The model discretizes the vine robot into 30 rigid bodies. Contact points are shown in red.}
    \label{fig:circle}
\end{figure}

To characterize the runtime performance of the simulator, we consider a vine robot growing at a constant growth rate in an environment with a single, circular object. Fig.\ \ref{fig:circle} shows an example of a 4-second trajectory in this environment, with a discretization of the vine robot into 30 rigid bodies. In this example, the timestep $\Delta t$ was 10 ms, and the average runtime per timestep was 3.0 ms. 

\begin{figure}[bp]
    \centering
    \includegraphics[]{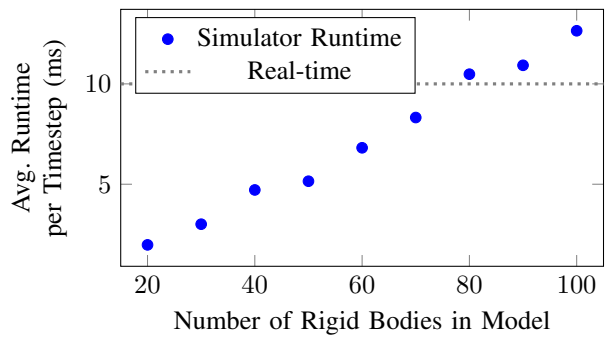}  
    \caption{Runtime results for different discretizations of the vine robot. For each model, we computed a 4-second trajectory, and recorded the average runtime per timestep. The plot shows a linear relationship between runtime and model size. The length of each timestep was 10 ms, demonstrating that the simulator is able to generate trajectories in real time for upwards of models with 70 bodies.}
    \label{fig:timing}
\end{figure}

We repeated this procedure for different discretizations of the vine robot, which correspond to different numbers of bodies in the model, and plotted the average runtime per timestep for each model in Fig.\ \ref{fig:timing}. These results show that (1) there is a linear relationship between model size and simulator runtime, and (2) the simulator runs in real time for models with upwards of 70 bodies. 

\subsection{Accuracy}\label{subsec:accuracy}

To characterize the accuracy of behaviors generated by the simulator framework, we present two examples of fitting model parameters using the optimization problem defined in Section \ref{sec:fitting}. In these examples, we specify a 10-body model and fit scalar values for the stiffness and damping coefficients. In choosing to fit scalar values, we are assuming that stiffness and damping are the same at each joint, and that these torques are linear in behavior. However, the fitting procedure can accommodate more general stiffness and damping models as well. Both examples use Interior Point OPTimizer (IPOPT) as the optimization solver \cite{Wachter2006}.

\subsubsection{Oscillation}

\begin{figure}[bp]
\centering
\includegraphics[]{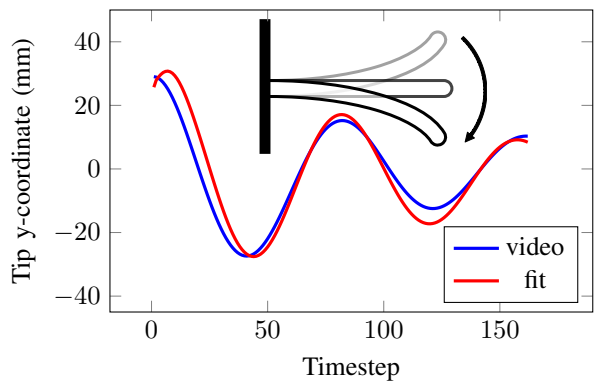} 
\caption{Results from the fitting method presented in this paper for a cantilevered robot oscillating upon release from an initial displacement in the y-axis. The plot compares the y-coordinates of the robot tip in the trajectory from the reference against that from the fit model. The mean absolute error is 3.3 mm. Overlaid on the plot is an illustration of the oscillating motion captured in the trajectories. }
\label{fig:oscillation}
\end{figure}

As a baseline example, we consider the scenario of a cantilevered robot oscillating upon release from an initial displacement (Fig.\ \ref{fig:oscillation}). In this example, there are no objects in the environment, and there is no growth of the robot over time. The reference trajectory used in the fitting procedure came from experimental high-speed video of an inflated beam marked with fiducials. We extracted a section of the video corresponding to two oscillations of the beam. The full video showed third-order dynamics that are noticeable over a longer time horizon, and we plan to address modeling this more complex behavior in future work. 

The results of the fitting procedure are plotted in Fig.\ \ref{fig:oscillation}. As a metric for accuracy, we compared the y-coordinate trajectory of the robot tip between the fit model and the real robot. The mean absolute error of the y-coordinate was 3.3 mm. Our results validate the ability of the simulator framework to capture bending modes of the vine robot. 

\subsubsection{Growth Into Wall}

\begin{figure}[tp]
\centering
\vspace{.2mm}
\includegraphics[]{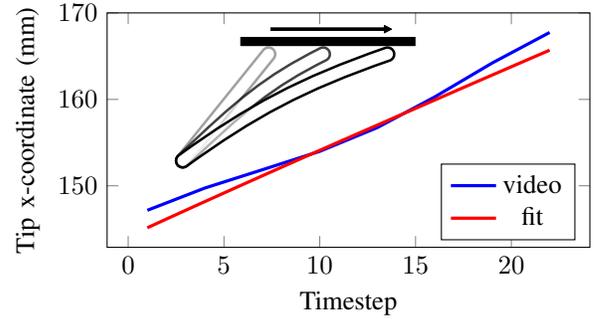}
\caption{Results from the fitting method presented in this paper for a vine robot growing into a wall. The plot compares the x-coordinates of the robot tip in the trajectory from the reference against that from the fit model. The mean absolute error is 1.1 mm. Overlaid on the plot is an illustration of the robot growth and contact interaction captured in the trajectories. }
\label{fig:wall}
\end{figure}

As an example of environment interaction, we consider the scenario of a vine robot that grows and comes in contact with a wall, causing the distal endpoint to slide along the wall as the robot bends and continues growing. We illustrate the contact interaction in Fig.\ \ref{fig:wall}. This fitting example incorporates robot growth and contact with its environment. The reference trajectory used in the fitting procedure came from a video of the scenario described above. We extracted a section of the video during which the robot was in contact with the wall.

The results of the fitting procedure are plotted in Fig.\ ~\ref{fig:wall}. As a metric for accuracy, we compared the x-coordinate trajectory of the robot tip between the fit model and the real robot. The mean absolute error of the x-coordinate was 1.1~mm. Our results validate the ability of the fitting procedure to optimize parameters in more complicated scenarios involving contact and growth. 

\subsection{Cluttered Environment Demonstration}\label{subsec:cluttered}

To demonstrate that the simulator can generate realistic behaviors with multiple contact interactions, we present a simulation of a robot growing in a cluttered environment. We based the environment structure on an existing video to show the qualitative agreement between the simulated and real trajectories. For the simulation, we used a model with 40 bodies and generated a 3-second trajectory. The results are overlaid on the original video in Fig.\ \ref{fig:Overview}. This example features simultaneous contact interactions at multiple points along the vine robot as well as navigation through passageways.

\section{Conclusion}

In this work, we presented a dynamics simulator for soft growing robots that generates realistic behaviors, handles environment interactions, runs in real time, and is amenable to motion planning. We designed a framework that uses a simplified multi-link model and incorporates existing methods for rigid-body dynamics with contact. We provided a fitting procedure that bridges the sim-to-real gap. Finally, we characterized the performance of the simulator framework in terms of runtime, accuracy, and capability. The computational efficiency and amenability to motion planning of our high-speed simulator validates its potential for use in the control of soft robots. Our implementation of the simulator framework is available at \url{ https://github.com/charm-lab/Vine_Simulator}.

We are interested in multiple areas of future work. First, because vine robots grow by adding new material at the tip, we can improve our rigid-body model by updating the mass matrix as the simulated robot grows. That is, a time-varying mass matrix would more accurately model the time-varying mass of a growing vine robot. Second, fitting more complex stiffness models would enable the simulator to capture the non-linear bending behaviors of the vine robot. Third, there are a number of performance optimization strategies that would greatly improve runtime performance. Fourth, the current model only accounts for actuation to grow, and a direction for future work is to account for bending actuation models. Finally, we are interested in integrating the presented simulator framework with motion planning strategies.

\section*{Acknowledgment}

The authors thank Joseph Greer, Laura Blumenschein, James Pillot IV, and Lauren Pitzer for contributing vine robot video data.

\bibliographystyle{IEEEtran}
\IEEEtriggeratref{11}
\bibliography{library}

\end{document}